# Exploit imaging through opaque wall via deep learning

Meng Lyu[1,2], Hao Wang[1,2], Guowei Li[1,2], and Guohai Situ[1,2,*]

[1]Shanghai Institute of Optics and Fine Mechanics, Chinese Academy of Sciences, Shanghai 201800, China

[2]University of the Chinese Academy of Sciences, Beijing 100049, China

*ghsitu@siom.ac.cn

**Abstract**

Imaging through scattering media is encountered in many disciplines or sciences, ranging from biology, mesescopic physics and astronomy. But it is still a big challenge because light suffers from multiple scattering is such media and can be totally decorrelated. Here, we propose a deep-learning-based method that can retrieve the image of a target behind a thick scattering medium. The method uses a trained deep neural network to fit the way of mapping of objects at one side of a thick scattering medium to the corresponding speckle patterns observed at the other side. For demonstration, we retrieve the images of a set of objects hidden behind a 3mm thick white polystyrene slab, the optical depth of which is 13.4 times of the scattering mean free path. Our work opens up a new way to tackle the longstanding challenge by using the technique of deep learning.

Conventional optical imaging theory relies on the assumption that the propagation direction of every wavelet emitting from the target to the entrance pupil of the optical imaging system does not change[1]. However, in applications ranging from *in situ* biomedical inspection in life science[2] to the free space optical communication in foggy and hazy environment[3], this assumption does not hold as scattering media exist in between the target and the optical system, which distorts the propagation direction of each wavelet in a random manner. As a result, no image about the target can be formed. Imaging through scattering media has been one of the main challenges in optics and received intensive interests in the history. Back to the 1960's, people have demonstrated that one can compensate the distortion introduced by turbulence or ground glass diffuser by using holography[4,5]. Holography has been extended to produce a phase-conjugated beam of the scattered one[6] so that the distortion can be removed when the beam propagates back through the scattering medium once again, and to separate the ballistic photons by taking the advantage of coherence gating[7]. Furthermore, it has also been demonstrated that holography has the potential to exploit three-dimensional object behind a thin diffuser[8]. Vellekoop and Mosk extended the concept of phase conjugation and proposed to optimize the wavefront at the input of a disordered slab so as to maximize the intensity at a given position at the other side of the slab[9,10]. Wavefront shaping technique does not require interferometric detection, raster-scanning or off-line reconstruction, and can be used for imaging through 'wall' and around a corner under coherent[11] and incoherent[12] illumination. As this technique allows one to manipulate light at a mesoscopic level in the scattering media, it can be used to measure the transmission matrix (TM)[13] and to exploit the transmission of image through opaque media[14]. The measurements of a scattering matrix enable a complete description and manipulation of multiple scattering in scattering media, these matrices have vast numbers of elements owing to the high degree of freedom[15]. Nowadays, no phase modulation device can perform precise complex-valued wavefront shaping or TM measurement at the wavelength scale, imposing limit to these powerful techniques. Memory effect[16], on the other hand, promises non-invasive[17] and single-shot imaging through scattering layer, under incoherent[18] and coherent illumination[19]. However, the correlation length of the light field that passes through the scattering slab drops very quickly to the wavelength scale when the slab becomes thicker[16].

Here, we exploit the technique of deep learning[20, 21] for imaging through thick scattering media. The proposed approach fits the inverse process of such an imaging system by a deep neural network (DNN). By using a well-trained DNN model, the image of an object hidden behind the scattering slab can be retrieved from the speckle intensity recorded by a camera. In the conceptual demonstration, we can retrieve different images hidden behind a 3mm thick white polystyrene slab from a single shot measurement of the corresponding speckle pattern with the DNN. The imaging depth is about 13.4 $l_s$, where $l_s$ is the scattering mean free path.

The optical setup used in our experiments is schematically illustrated in Fig. 1(a). A collimated and expanded laser beam was incident on an amplitude-only liquid crystal SLM after passing through the linear polarizer P1 and variable iris 1. Target images are displayed on the SLM to modulate the incident beam. The reflected beam went through the second variable iris 2, the second linear polarizer P2 and passed through a 3 mm thick scattering slab successively. The polarization states of P1 and P2 are set to be perpendicular to each other. In order to visually display the scattering of the screen, we captured the images at the front and back surface of the slab and show them in Fig. 1(b) and 1(c). It is clearly seen that the image is highly visible at the front surface and becomes severally diffusive after it passes through the slab, containing no visible information about the target. This is reasonable as we measured the scattering mean free path of the slab, $l_s$, and it is 224 μm as we measured. The optical thickness of the scattering slab is then about 13.4 $l_s$, so that the photons randomly walk beyond the diffusive transport regime[2].

Nevertheless, the speckle image $I$ that is captured by the camera is related to the target $O$ through

$$I = f(O) \qquad (1)$$

Our purpose is then to retrieve the target image $O$ from the speckle captured by the camera, and it is stated mathematically as

$$O = f^{-1}(I) \qquad (2)$$

Note that, within the isoplanatic angle, Eq. (1) reduces to the convolution $I(\mathbf{r}) = \int d\mathbf{r}^2 O(\mathbf{r}')h(\mathbf{r}-\mathbf{r}')$, where $h$ is the point spread function of the system. With the presence of thick scattering media, this isoplanatic angle is zero[22] so that the convolution simplification only holds for a point source. In general, $f$ is a multi-input-multi-output function, and is too complex to measure precisely. As a consequence, it is infeasible to solve Eq. (2) directly for the target image. Indeed, in a more generalized form, various methods have been proposed to measure the transmission matrix[23]. Essentially, the idea is to send a set of input spatial modes into the medium, and holographically record the transmitted amplitude corresponding to each of these modes, and then determine directly from these input-output measurements the transmission matrix elements. However, due to the limited pixel number of currently available SLMs, one needs to significantly demagnify the SLM phase pattern using a microscopic objective with large numerical aperture so that only a small number of modes of the scattering media can be explored.

The machine-learning-based method we propose here does not rely on the mesoscopic physical detail of the transmission matrix. Alternatively, it needs a large set of input-output measurements to train the DNN so as to learn how the output is related to the input. In theory, this works very well provided that the network has at least one hidden layer and sufficient number of hidden neurons[24,25]. Thus, in this work, we used a DNN model with two reshaping layers, four hidden layers and one output layer. The layout of the network is schematically shown in Fig. 2. To train the DNN, we used the images of 3990 handwritten digits downloaded from the MNIST handwritten digit database[26] as the training targets, and sequentially displayed them on the amplitude-only SLM shown in Fig. 1. The camera then captured the corresponding speckle images. The set of 3990 input-output data was fed into the DNN model to optimize the connection weights of every neurons (see Methods for details). After 200 epochs, the mean-squared error (MSE) between the network output and the nominal appearance of the handwritten digits drops down to 0.03, and becomes steady. Then the network is trained.

To test the performance, we randomly selected the images of five other digits from the MNIST database and sent them to the SLM, and obtained five speckle images [see, first row in Fig. 3]. The retrieved target images are shown in the second row in Fig. 3(a). In comparison to the ground truth images (shown in the third row), all the visible features, in particular, the edges, of the targets have been retrieved, with the MSE down to 0.03. The edges are more robustly retrieved because they are first built from the speckle pixels in the neural network, and subsequently formed the retrieved images[20,21]. Unlike the images reconstructed using the phase conjugation[6], or wavefront shaping[11], or transmission matrix[14], the background of the images retrieved using the DNN is very clean. However, due to the challenges to optimize the network and the overfitting issue[20], dark tripes or spots appear in the retrieved images. Further considerations need to take to improve the images.

As the scattering slab we used in our experiment has an optical thickness of 13.4 $l_s$, the photons observed by the camera are in the diffusive regime. The memory effect range becomes too small to measure in this case. As a consequence, the techniques developed by Bertolotti[17], Katz *et al.*[18] and Li *et al.*[19] that are based on memory effect would fail. This is demonstrated by the experimental results plotted in the bottom row of Fig. 3. The results suggest that the proposed deep-learning-based method can retrieve the image of targets far beyond the memory effect range.

The DNN trained by the digit images alone is good enough to retrieve the images of handwriting English letters from the corresponding speckle intensities. The experimental results are plotted in Fig. 4. As the DNN was trained using the image set of digits, it inclines to build the same type of image from the speckle of the testing images. This is evident by looking at the retrieved image of letter 'z', which looks like the digit '2'. This issue can be resolved by using the set of training data of various types.

As described above, the proposed deep-learning-based method uses arbitrary images to train unsupervisely the DNN, rather than using a set of spatial modes[13], or of plane waves with different $k$[23] as in the TM methods. And it does not require to measuring the complex amplitude, as it just needs the input-output intensities to train. Furthermore, it does not need to use any microscopic objective to demagnify the patterns that are displayed on the SLM to fit the size of the eigenchannel[9]. Thus, the trained DNN does not reveal any physical detail about the transmission matrix of the slab. Nevertheless, it shows promise to retrieve images of the type other than those that were used to train the network. And this opens up new possibility to explore optical imaging through more general complex systems.

**Methods**

**Optical experimental setup**

In the optical experiments, the laser we used was a linearly polarized optically pumped semiconductor lasers (Verdi G2 SLM, Coherent, Inc.) irradiating at 532 ± 2 nm. The amplitude-only SLM is a Holoeye Pluto-Vis with the maximum pixel count of 1920×1080 and the pixel size of 8.0 μm×8.0 μm. The images size of handwritten digits from the MNIST database is 28×28 pixels. We magnified these images 18 times and displayed them on the SLM after zero padding to the size of 1920 × 1080 pixels. As a result, the physical size of digit images on the SLM is about 2.34mm × 2.34mm. The scattering slab was a 3 mm thick white polystyrene (EDU-VS1/M, Thorlabs, Inc.), the scattering mean free path of which was about 224 μm as we measured. The camera used in the experiments is an Andor Zyla 4.2 PLUS sCMOS with a 4.2 megapixel sensor. But we only activated the central 64×64 pixels in the experiments to capture the speckle intensity. Variable iris 1 in Fig. 1 controls the size of laser beam, and variable iris 2 was used to filter out the high diffraction orders generated by the SLM. $d_1$ represents the distance between the SLM and the slab, and $d_2$ represents the distance between the slab and the sensor of the camera. In our experiments, $d_1$ = 290mm and $d_2$ = 35mm.

**The DNN model**

The DNN model has two reshaping layers, four hidden layers and one output layer as shown in Fig. 2. Reshaping layer 1 reshapes the 64 × 64-pixel input speckle image into a 1 × 4096 vector. Hidden layer 1 has 8192 neurons, hidden layer 2 has 4096 neurons, hidden layer 3 has 2048 neurons, hidden layer 4 has 16384 neurons and output layer has 784 neurons. The activate function of these neurons is rectified linear units (ReLU) which allow for faster and effective training of deep neural architectures on large and complex datasets compared with the activate functions in traditional neural network such as sigmoid function[20,27]. Reshaping layer 2 reshapes the 1 × 784-pixel vector coming from the hidden layer into a 28 × 28 image. The loss function and optimization in the DNN model is MSE and stochastic gradient descent (SGD). The program was implemented using Python version 3.5 and the DNN was implemented using Keras framework based on TensorFlow framework, and the implementation was GPU-accelerated with a NVIDIA Tesla K20c.

**References**

[1]. Goodman, J. W. Introduction to Fourier optics (Roberts and Company Publishers, 2005).

[2]. Ntziachristos, V. Going deeper than microscopy: the optical imaging frontier in biology, Nat. Methods 7, 603–614 (2010).

[3]. Ishimaru, A. Wave Propagation and Scattering in Random Media, (Academic Press, 1978).

[4]. Goodman, J., Huntley Jr, W., Jackson, D. & Lehmann, M. Wavefront-reconstruction imaging through random media. Appl. Phys. Lett. 8, 311–313 (1966).

[5]. Kogelnik, H. & Pennington, K. Holographic imaging through a random medium. J. Opt. Soc. Am. 58, 273–274 (1968).

[6]. Yaqoob, Z., Psaltis, D., Feld, M. S. & Yang, C. Optical phase conjugation for turbidity suppression in biological samples. Nat. Photon. 2, 110–115 (2008).

[7]. Zhang, Y., Situ, G., Wang, D., Pedrini, G., Javidi, B., & Osten, W. Application of short-coherence lensless Fourier-transform digital holography in imaging through diffusive medium. Opt. Commun. 286, 56–59 (2013).

[8]. Singh, A. K., Naik N. D., Pedrini, G., Takeda, M., & Osten, W. Exploiting scattering media for exploring 3D objects. Light Sci. Appl. 6, e16219 (2017).

[9]. Vellekoop, I. M. & Mosk, A. Focusing coherent light through opaque strongly scattering media. Opt. Lett. 32, 2309–2311 (2007).

[10]. Vellekoop, I. M., Lagendijk, A. & Mosk, A. Exploiting disorder for perfect focusing. Nat. Photon. 4, 320–322 (2010).

[11]. Mosk, A. P., Lagendijk, A., Lerosey, G. & Fink, M. Controlling waves in space and time for imaging and focusing in complex media. Nature Photon. 6, 283–292 (2012).

[12]. Katz, O., Small, E. & Silberberg, Y. Looking around corners and through thin turbid layers in real time with scattered incoherent light. Nat. Photon. 6, 549–553 (2012).

[13]. Popoff, S. et al. Measuring the transmission matrix in optics: an approach to the study and control of light propagation in disordered media. Phys. Rev. Lett. 104, 100601 (2010).

[14]. Popoff, S., Lerosey, G., Fink, M., Boccara, A. C. & Gigan, S. Image transmission through an opaque material. Nat. Commun. 1, 81 (2010).

[15]. Yu, H. et al. Recent advances in wavefront shaping techniques for biomedical applications. Curr. Appl. Phys. 15, 632–641 (2015).

[16]. Freund, I., Rosenbluh, M., & Feng, S. Memory effect in propagation of optical waves through disordered media. Phys. Rev. Lett. 61, 2328–2332 (1988).

[17]. Bertolotti, J. et al. Non-invasive imaging through opaque scattering layers. Nature 491, 232–234 (2012).

[18]. Katz, O., Heidmann, P., Fink, M. & Gigan, S. Non-invasive single-shot imaging through scattering layers and around corners via speckle correlations. Nat. Photon. 8, 784–790 (2014).

[19]. Li, G., Yang, W., Li, D., & Situ, G. Cyphertext-only attack on the double random-phase encryption: experimental demonstration. Opt. Express, 25, 8690-8697 (2017).

[20]. Goodfellow, I., Bengio, Y. & Courville, A. Deep learning (MIT press, 2016).

[21]. LeCun, Y., Bengio, Y. & Hinton, G. Deep learning. Nature 521, 436–444 (2015).

[22]. Freund, I. Image reconstruction through multiple scattering media. Opt. Commun. 86, 216-227 (1991).

[23]. Rotter, S. & Gigan, S. Light field in complex media: mesoscopic scattering meets wave control. Rev. Mod. Phys. 89, 015005 (2017).

[24]. Hornik, K., Stinchcombe, M. & White, H. Multilayer feed forward networks are universal approximators. Neural Netw. 2, 359–366 (1989).

[25]. Cybenko, G. Approximation by superpositions of a sigmoidal function. Math. Control Signals Syst. 2, 303–314 (1989).

[26]. Deng, L. The MNIST database of handwritten digit images for machine learning research. IEEE Signal Process. Mag. 29, 141–142 (2012).

[27]. Nair, V. & Hinton, G. E. Rectified linear units improve restricted Boltzmann machines. In Proc. Int. Conf. Mach. Learn. (IEEE, 2010), 807–814 (2010).

**Funding**

National Natural Science Foundation of China (61377005), and the Chinese Academy of Sciences (QYZDB-SSW-JSC002).

**Competing Interests:** The authors declare that they have no competing interests.

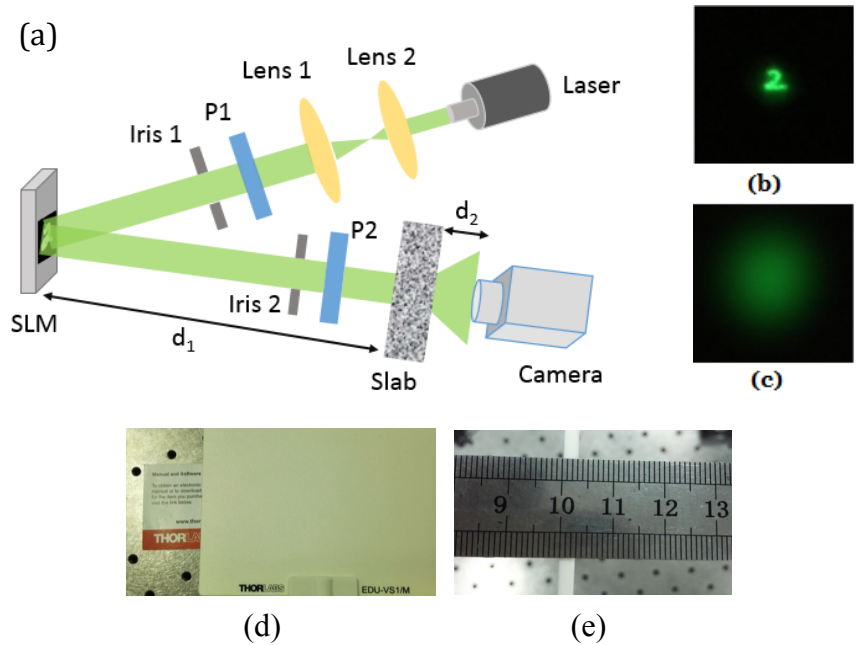

Figure 1. (a) Experimental setup for imaging through scattering media, SLM represents an amplitude-only spatial light modulator, P1 and P2 are linear polarizers and the slab is a 3mm thick white polystyrene. The image captured at the front (b) and back (c) surface of the scattering medium. The side view (d) and top view (e) of the polystyrene.

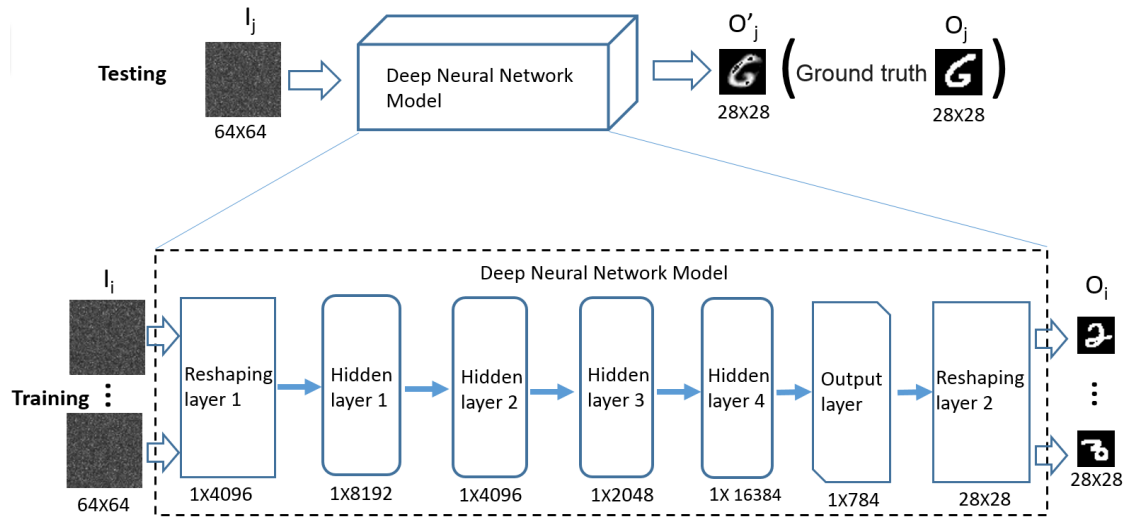

Figure 2. The diagram of the DNN model. I$_i$ and O$_i$ are the training input and output of the DNN model, respectively. Ij and O′ are the testing input and output of the DNN model, respectively. O j is the ground truth.

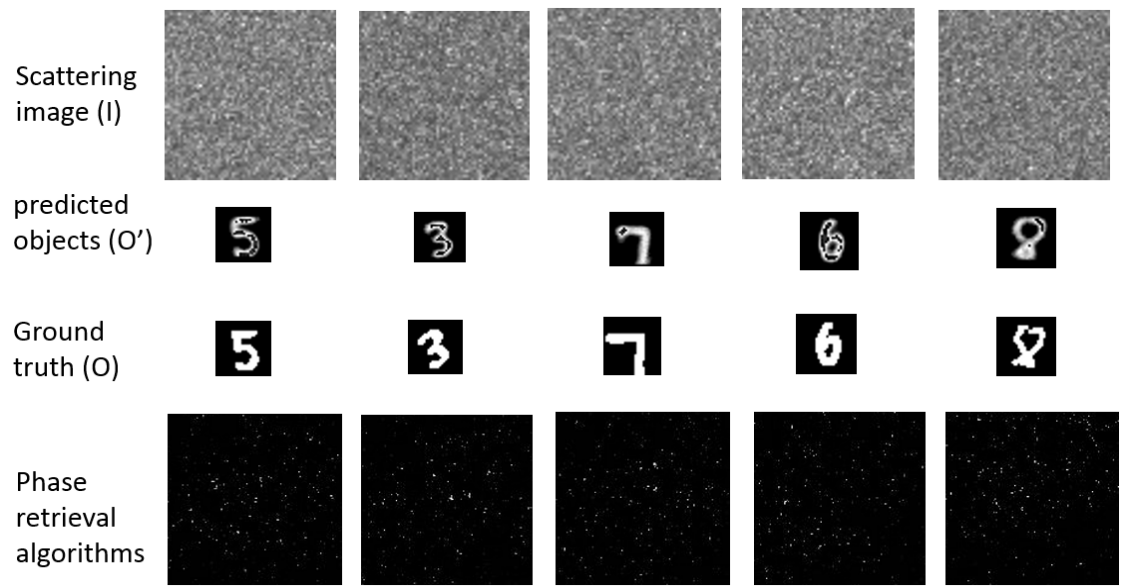

Figure 3. The results of the digits. The images in the first row are the speckle images recorded by the camera which are also inputs to the DNN model, the second row shows the predicted objects, the third row are the ground truth and the fourth row are the reconstructed images by retrieve algorithm.

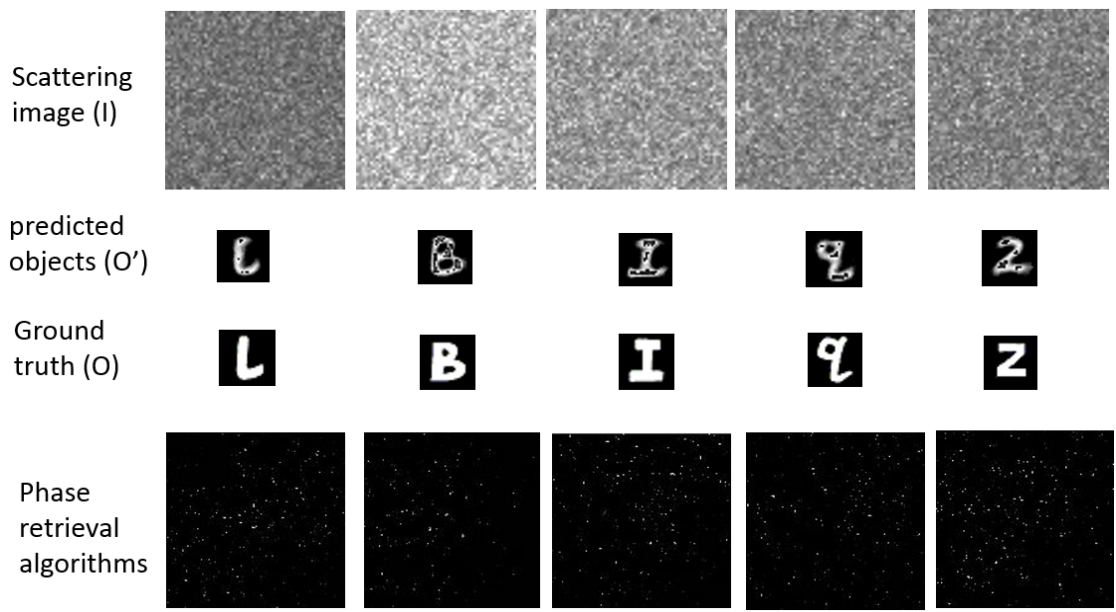

Figure 4. The results of the English letters. The images in the first row are the speckle images recorded by the camera which are also inputs to the DNN model, the second row shows the predicted objects, the third row are the ground truth and the fourth row are the reconstructed images by retrieve algorithm.